\pdfoutput=1

\documentclass[11pt]{article}

\usepackage[]{acl}

\usepackage{times}
\usepackage{latexsym}

\usepackage[T1]{fontenc}

\usepackage[utf8]{inputenc}

\usepackage{microtype}

\usepackage{inconsolata}

\usepackage{graphicx}
\usepackage{tikz} 
\usepackage{pgfplots} 
\pgfplotsset{compat=1.18} 
\usepgfplotslibrary{colormaps}
\usepackage{bbm}

\usepackage{multirow}

\usepackage{adjustbox}
\usepackage{booktabs}
\usepackage[most]{tcolorbox}
\usepackage{listings}
\usepackage{colortbl}
\usepackage{array}
\usepackage{caption}
\usepackage{subcaption}
\usepackage{fontawesome5}
\usepackage{svg}
\definecolor{lightgray}{rgb}{0.95,0.95,0.95}
\newcommand{\dataset}{\textsc{ConFETTI}}
\title{\dataset: Conversational Function-Calling Evaluation \\Through Turn-Level Interactions}

\author{Tamer Alkhouli, Katerina Margatina, James Gung,  \\ \textbf{Raphael Shu, Claudia Zaghi, Monica Sunkara, Yi Zhang} \\
    \{alkhouli, katemarg, gungj, zhongzhu, sunkaral, yizhngn\}@amazon.com \\
     {\raisebox{-0.5ex}{\makebox[17pt][l]{\Large \faAws }} AI Labs} 
    }
  
\usepackage{pgfplotstable}

\begin{document}
\maketitle
\begin{abstract}

\end{abstract}
We introduce \textbf{Con}versational \textbf{F}unction-Calling \textbf{E}valuation \textbf{T}hrough \textbf{T}urn-Level \textbf{I}nteractions (\textsc{ConFETTI}), a conversational  benchmark\footnote{\url{https://github.com/amazon-science/confetti}} designed to evaluate the function-calling capabilities and response quality of large language models (LLMs). Current benchmarks lack comprehensive assessment of LLMs in complex conversational scenarios. \dataset{} addresses this gap through $109$ human-simulated conversations, comprising $313$ user turns and covering $86$ APIs. These conversations explicitly target various conversational complexities, such as follow-ups, goal correction and switching, ambiguous and implicit goals. We perform off-policy turn-level evaluation using this benchmark targeting function-calling. Our benchmark also incorporates dialog act annotations to assess agent responses.  We evaluate a series of state-of-the-art LLMs and analyze their performance with respect to the number of available APIs, conversation lengths, and chained function calling. Our results reveal that while some models are able to handle long conversations, and leverage more than 20+ APIs successfully, other models struggle with longer context or when increasing the number of APIs. We also report that the performance on chained function-calls is severely limited across the models. Overall, the top performing models on \dataset{} are Nova Pro (40.01\%), Claude Sonnet v3.5 (35.46\%) and Llama 3.1 405B (33.19\%) followed by command-r-plus (31.18\%) and Mistral-Large-2407 (30.07\%).
\section{Introduction}

Function-calling has emerged recently as one of the notable capabilities of large language models (LLMs)~\cite{schick2023toolformer, yao2023react, Qin:2024, patil2024gorilla}. To generate a function call, LLMs are provided with a list of available functions, including function descriptions, schema, parameter names and types. 
When provided with natural language input, LLMs can generate appropriate function calls to fulfill user requests. Our work specifically examines interactive conversational scenarios where the user and the system engage in dynamic dialogue. These conversations evolve through multiple turns, including follow-up questions, requests for clarification, and confirmation checks, creating what we define as a conversation trajectory. 
\begin{figure*}[t]
    \centering
    \resizebox{\textwidth}{!}{
    \includegraphics{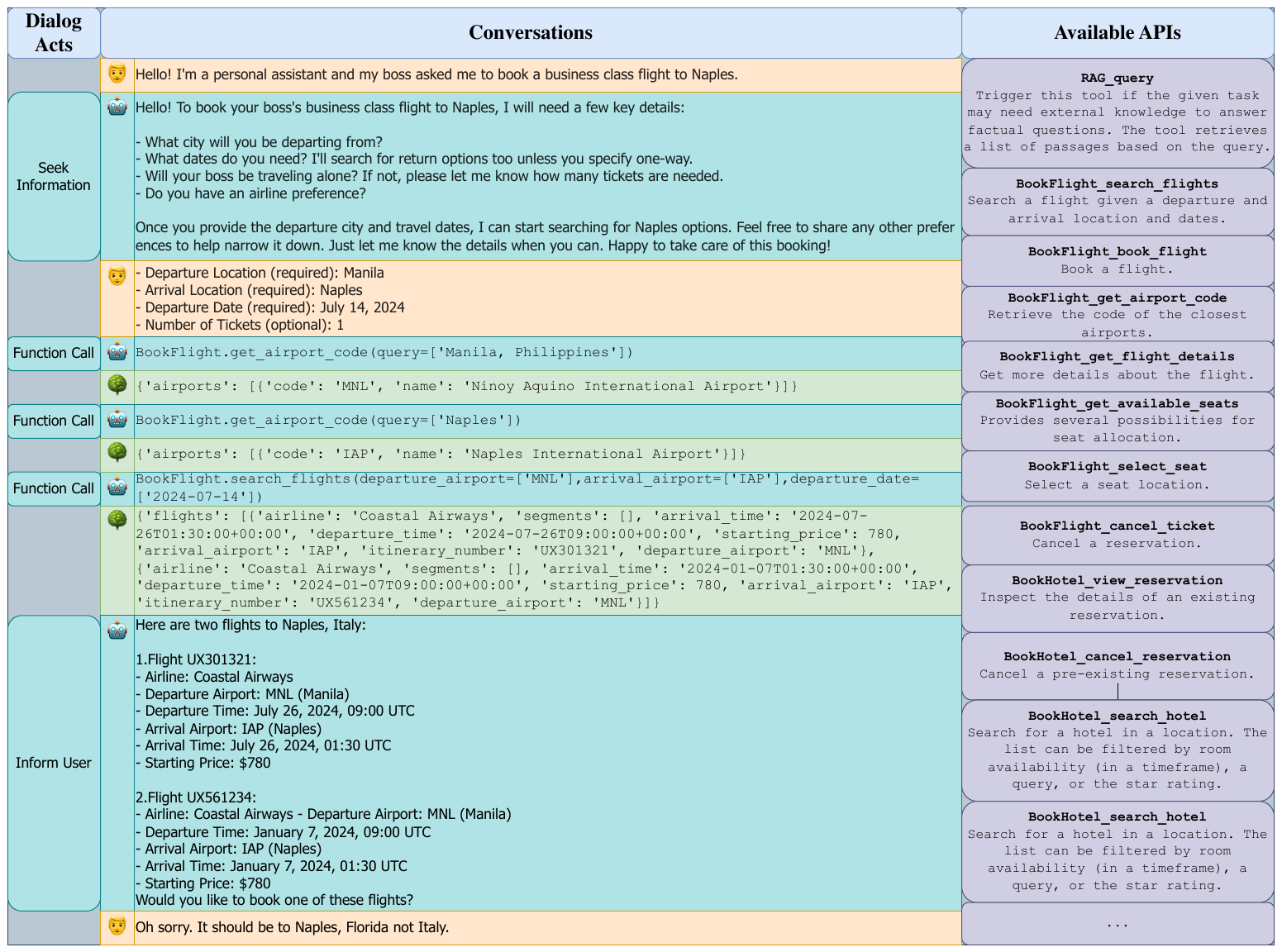}
    }
    \caption{\dataset{} comprises multi-turn conversations between a user and an agent. The agent has access to a set of APIs (tools) to assist the user with their request. The agent can invoke actions and receive observations from the environment. Each agent turn is annotated with a dialog act tag used to evaluate response quality. The figure illustrates the flow of a conversation, including user inputs, agent responses, API calls, and dialog act annotations.}
    \label{fig:confetti_details}
\end{figure*}

There can be multiple conversation trajectories that achieve the same user goals. This variation adds complexity to evaluating models in conversational multi-turn sessions. There are several ways to perform automatic evaluation in conversational settings. 
First, the evaluation setup can use a user simulator to automate interaction with the model dynamically, where user utterances are simulated online \cite{Asri2016ASM,lu2024toolsandbox,sek2024usersim,yoon2024evaluatinglargelanguagemodels}. This adds flexibility to the evaluation to accommodate potential variations in the conversation trajectories. However, the evaluation will be subject to potential errors stemming from the user simulator itself, which can obscure the actual performance of the model. 
Alternatively, user utterances can be determined prior to the conversation session to achieve one or more user goals \cite{mao2024bfcl}. This, however, assumes the trajectory will follow a certain path, with deviations ending up being penalized.
A third approach is to evaluate conversations on the turn-level. This is done in offline mode where the history is set to the ground-truth trajectory up to the current turn in the conversation. Although it is done offline, it isolates the behavior of the model at each turn, and allows for a direct comparison between predictions and ground-truth labels. In this work, we follow the third approach.  

Our contributions are as follows:
\begin{itemize}
    \item We introduce a \textbf{conversational function-calling benchmark} targeting conversation phenomena such as follow-ups, goal correction and switching,
ambiguous and implicit goals, over- and underfilling of parameters, etc.
Figure~\ref{fig:confetti_details} illustrates an example.
    \item We propose \textbf{response quality evaluation} to evaluate both function-calling and non-function calling responses via dialog act classification,  categorizing responses as seeking information, providing information, rejecting requests or calling functions.
    \item We  \textbf{compare state-of-the-art LLMs}, from various model providers, showcasing their potentials and gaps on conversational function-calling.
\end{itemize}

\section{Related Work}
ToolSandbox \cite{lu2024toolsandbox} offers an  end-to-end evaluation approach, where the authors employ a dynamic user simulator with intermediate milestones to evaluate trajectories. The dataset lacks in domain diversity and complexity, focusing on productivity tools. API-Bank \cite{li2023apibank} covers multi-turn dialog and performs API executions. Tooltalk \cite{farn2023tooltalk} performs turn-level evaluations, and ToolEmu \cite{ruan2024identifyingriskslmagents} mocks API responses, similar to our approach. AgentsBench \cite{liu2023agentbench} offers open-ended dialog evaluation of LLMs as Agents for Code, Gaming, and Web environments. PlatoLM \cite{kong-etal-2024-platolm} proposes to train a user simulator and synthesize conversations as training data.

BFCL v3 \cite{mao2024bfcl} targets evaluating function-calling capabilities in single- and multi-turn settings. Its multi-turn dataset is closely related to our work. There are however several differences. While the BFCL multi-turn data is model-generated and validated by a human, our data is completely generated by human annotators. This avoids potential bias towards certain model families. Second, our collection targeted multiple conversational complexities such as confirmations, follow-ups, and elicitation as part of the natural flow of the conversations. BFCL offers a subset where parameters are missing to induce elicitation. Third, we include $3$-$25$ functions to predict from at each turn, aiming to reveal model capabilities under varying number of functions provided.  

There have been numerous studies on dialog act modeling \cite{stolcke-da} and classification \cite{ahmadvand-da,kimd-da,liu-da,Chakravarty2019DialogAC,Duran_Battle_Smith_2023}. In this work, we use an LLM-based approach to classify agent dialog acts into 5 labels. 

LLM as a judge \cite{wang-etal-2023-chatgpt,pan-etal-2024-human,liu-etal-2023-g,Zheng2023JudgingLW} has been explored recently. In this work, we leverage LLMs as a means to detect parameter hallucinations.

\section{\dataset{}}
\subsection{Data Collection}
\paragraph{Complexities} Following ~\citet{gung-etal-2023-natcs}, to ensure that the conversations in our benchmark cover a broad range of interaction patterns, we define a set of \textbf{complexities} that may be observed in real world interaction with conversational function calling systems.
These complexities, defined in Table~\ref{tab:conversation_complexities}, are used as an input to the conversation authoring process with a target distribution enforced by requiring at least one complexity for each conversation.
For example, if \textsc{ExceptionInExecution} is assigned as a required complexity in a conversation, at least one function call must have an error response for it to be considered valid.
Annotators are also asked to label the complexities that appear in their conversations for tracking purposes.

\paragraph{APIs} We curated a broad range of APIs across application areas such as issue tracking, travel booking, human resources, and meetings management.
Each API contains a set of function definitions with schemas defining valid inputs and outputs for each function.
APIs do not have explicit implementations in our benchmark, but instead have fixed outputs defined in each conversation.

\paragraph{Scenario Generation} Each conversation is seeded with a set of required and optional APIs, a required complexity, a reference time the conversation is to take place, and a minimum number of turns.
Based on these inputs, a scenario is defined that describes one or more user goals for the conversation in natural language.
Authoring valid scenarios requires defining a realistic request that can be resolved given the available APIs and integrating the assigned complexity.
Due to the difficulty of this task, we separated the authoring of scenarios from conversations, only allowing conversations to be instantiated for scenarios after they were reviewed by trusted annotators.

\rowcolors{2}{gray!25}{white} 
\begin{table*}[!h]
\small
\centering 
\resizebox{0.99\textwidth}{!}{ 
\begin{tabular}{lcp{0.6\textwidth}} 
\toprule 
\textsc{Complexity} & \textsc{\# dialogs} & \textsc{Description} \\
\hline
\textsc{ExceptionInExecution} & 5 & Errors or exceptions that occur during the execution of an action \\ 
\textsc{FailedConversation} & 5 & Interactions where the intended goal is not achieved \\

\textsc{Confirmation} & 6 & Requesting user approval before executing an action \\

\textsc{GoalSwitching} & 6 & When the user changes their objective during the conversation \\

\textsc{NoTargetComplexity} & 6 & Conversations without specific complexity requirements \\ 

\textsc{GoalCorrection} & 7 & Adjusting or refining the user's goal based on feedback \\ 

\textsc{GoalStacking} & 7 & Managing multiple user objectives simultaneously \\

\textsc{AmbiguousGoal} & 9 & When the user's intention is unclear and requires clarification \\ 

\textsc{FollowUpQuestion} & 10 & Additional queries or requests for information after the initial response \\ 

\textsc{ImplicitDescriptiveGoal} & 10 & The user describes a problem/background without directly stating their goal \\

\textsc{Overfill} & 11 & Providing more information than required for an action \\ 

\textsc{Underfill} & 11 & Missing required arguments or information for an action \\

\textsc{GoalNotSupported} & 15 & The user's request is not supported by the available tools or is out of scope \\
\bottomrule 
\end{tabular} } \caption{The distribution and description of the complexities covered in the \textsc{ConFETTI} benchmark.} \label{tab:conversation_complexities} \end{table*}

\paragraph{Conversation Generation} Conversation authors play the role of both the user and the agent, enacting the pre-defined scenario by entering user turns, agent turns, and function calls that meet the required complexities and align with the scenario description.
Authors use dynamically-generated forms in the annotation UI to create valid function calls, which are validated upon submission to ensure the authored values are  valid for the function input schema at hand.
Upon submission of a function call and selection of a desired response type (i.e. success code vs. error codes), the  return value is simulated using an internal model based on the function description, input/output schema, and input values.

\paragraph {Dialog Acts} Following conversation generation, we asked annotators to label agent turns using one or more of the following dialog acts:
\begin{itemize}
    \item \textbf{Agent seeking information (seek\_info)}: targets responses that elicit information from the user, whether it is intent elicitation, parameter elicitation, or asking for clarification or confirmation.
    \item \textbf{Agent informing user (inform)}: targets responses that provide information to the user, whether the information is general or specific to the function-calling results.
    \item \textbf{Agent rejecting request (reject)}: targets responses that reject the user request.
    \item \textbf{Other}: any response that does not belong to the previous dialog acts.
\end{itemize}

Note that the agent response can including multiple dialog acts. We ensure that the annotators try to assign a single label before opting for multiple labels.
\rowcolors{1}{white}{white} 

\subsection{Benchmark Design}
After the data collection, we convert the conversations into input/output pairs. We split the data into two benchmarks: \textbf{function-calling} benchmark, and \textbf{response quality} benchmark. The function-calling conversational benchmark is defined at the turn-level, where we truncate the conversations at each function-calling turn, and provide the preceding user turns, agent turns, function calls and function results as context. The output is set to be one or more function calls. We extract all function-calling turns from each conversation to construct the dataset. We also include the API schemas of the APIs the agent can choose from. The dataset has a total of $506$ examples. Table~\ref{tab:function_calling_data_stats} shows the data statistics.

Similarly, we define the response quality benchmark at the turn-level. However, instead of truncating conversations at function-calling turns, we extract all agent turns as output turns. The input is set to the preceding user turns, agent turns, and function calls and results, as in the conversational function-calling benchmark, and the API schema of available APIs to call is provided as well. However, the output turns can be function-calling turns or textual agent responses. We limit the number of function-calling turns to 200 to maintain label balance in this dataset. The statistics are given in Table~\ref{tab:da_data_stats}.

\begin{table}[!h]
\small
\centering
\resizebox{0.7\columnwidth}{!}{ 
\begin{tabular}{lc}
\toprule
\textsc{Dimension} &  \textsc{count}  \\
\hline
\# conversations	& 109 \\
\# APIs	& 86 \\
\# user turns  & 313 \\	
\hline
\# agent turns w/ 1 action	 & 220 \\
\# agent turns w/ 2 actions  & 46 \\
\# agent turns w/ 3+ actions  & 47	\\
\hline
Avg turns per conversation  & 8.8	\\
\# total turns & 506 \\
 \bottomrule
\end{tabular}
}
\caption{Function-calling benchmark statistics.} \label{tab:function_calling_data_stats}
\end{table}

\begin{table}[!h]
\small
\centering
\resizebox{0.7\columnwidth}{!}{ 
\begin{tabular}{lc}
\toprule
\textsc{Dimension} &  \textsc{count}  \\
\hline
\# turns calling functions & 200 \\
\# turns informing user & 331 \\
\# turns seeking information & 110 \\
\# turns rejecting requests & 32 \\
\# other/misc. turns & 25 \\
\hline 
\# total turns & 663 \\
 \bottomrule
\end{tabular}
}
\caption{Response quality benchmark statistics: note a turn can have one or more dialog act labels.} \label{tab:da_data_stats}
\end{table}
\section{Evaluation Metrics}
We leverage abstract syntax trees (AST) to evaluate function-calling turns similar to BFCL v2. For response quality, we resort to dialog acts.  
\subsection{Function-Calling Evaluation Metrics}
\label{sec:param_valid}
\paragraph{AST Soft}
AST evaluation parses the function name, parameter names and values as a tree, and matches each of the tree nodes to the ground-truth reference function call in a binary fashion.
Since the dataset includes many APIs that accept string-type input, we use soft scoring to calculating parameter value matching for string types. We use AlignScore \cite{zha2023alignscore} to calculate the matching accuracy for string parameter values. All non-string parameter values are evaluated in a binary fashion.

\paragraph{Parameter Hallucination}
We also perform an analysis on how often the models hallucinate parameters at the action calls. We use an off-the-shelf LLM to do hallucination detection. The prompt includes the conversation history, the predicted action call (or chain of action calls) and their schemas, and instruct the LLM judge to return a list of hallucinated parameters (if any) for each action call, along with a rationale. We use \texttt{gpt-4o-mini} as an LLM judge to avoid bias in the evaluation. The exact prompt is shown in the Appendix~\ref{sec:app_hallucination}.

Given a sequence of gold actions $\mathcal{A}_\text{g}$ and predicted actions $\tilde{\mathcal{A}}$ corresponding to the input context sequence $\mathcal{C}$, and denoting $\tilde{a_t}$ as the predicted action for the $t$-th example, we calculate the parameter validity rate on the parameter level using the following formula.
\begin{equation*}
   \sum_{(\tilde{a}_t, a_t, c_t) \in (\tilde{\mathcal{A}},\mathcal{A}_{g}, \mathcal{C})} \sum_{\theta \in \tilde{a}_t}  \frac{\mathbbm{1}(\text{$\tilde{a_t}$ is valid}) \times f(\theta ; c_t, a_t) }{|\{\hat{\theta} | \hat{\theta} \in \tilde{a}_t \, , \forall \tilde{a_t} \in \tilde{\mathcal{A}}\}|} 
\end{equation*}
Where $\mathbbm{1}$ is the indicator function. We define valid actions as predicted actions whose name matches the ground-truth action. $f(\theta; c_t, a_t) \in \{0,1\}$ is the binary outcome of the LLM classifier judging the validity of parameter $\theta$ given the context $c_t$ and gold action $a_t$. The denominator denotes all predicted parameters of valid predicted actions.

\subsection{Response Quality Evaluation via Dialog Acts}
Calculating response quality on the textual responses directly is challenging. While there are many ways to calculate similarity between two texts, we resort to a coarse-grained response evaluation via dialog acts. To this end, we leverage dialog act annotations provided by the annotators for each agent response in the ground-truth conversations.  In order to evaluate online agent responses, we use an LLM-based dialog act classifier that is prompted to generate one or more of the dialog acts for an agent response.

\section{Experiments}
We implemented the benchmark as an extension to BFCL v2.\footnote{\url{https://github.com/ShishirPatil/gorilla/tree/main/berkeley-function-call-leaderboard}} We compare Claude Sonnet v3.5, Sonnet v3, Haiku v3, Llama 3.1 70B and 405B, Llama 3 70B, Mistral large, and the Command-plus-r models. We distinguish between using the tool use option in the API vs using direct prompting. The difference is that in the tool use case, the API handles the function call parsing, while in the direct prompting case, the function schemas and parsing are handled explicitly by the user. We use Bedrock's Converse API to conduct these experiments\footnote{There are two modes of calling the API, with tool use or without. In the tool use case, the tool schemas are passed as an API argument, as opposed to concatenating the schema to the prompt directly. We experimented with both, where we use $\dagger$ to indicate model called using the tool use option. We report the best of the two modes for each model.}. We use the \textsc{AlignScore-base} model to calculate soft scores between string parameter values for the AST soft accuracy metric.\footnote{\url{https://github.com/yuh-zha/AlignScore}}. Table~\ref{tab:func_prompt} and Table~\ref{tab:response_prompt} in Appendix~\ref{sec:app_hallucination} indicate the prompts used to generate predictions for the function-calling and response quality benchmarks, respectively. All experiments are run with temperature $0$. 
\subsection{Function-calling Performance}\label{sec:func_call_eval_results}

Table~\ref{tab:model_accuracy_scores_by_provider} shows function-calling results measured in terms of AST soft accuracy across the different models. We observe Nova Pro (40.91\%) and Claude Sonnet v3.5 (35.46\%) among the top performing models, followed by Nova Micro (33.97\%), Llama 3.1 405B models (33.19\%), Llama 3.1 70B (31.29\%),  Command-r-plus (31.18\%)  and Mistral Large (30.07\%). Smaller models such as Nova Micro (33.97\%) and Haiku v3.5 (31.25\%) perform relatively well compared to the others.

\begin{table}[t!] \centering \resizebox{\columnwidth}{!}{ \begin{tabular}{llc}
\toprule 
\textsc{Model} & \textsc{Provider} & \textsc{AST Soft} (\%) \\
\hline
\texttt{Nova Pro} & Amazon & 40.91  \\
\texttt{Nova Micro} & Amazon & 33.97 \\
 \hline
\texttt{claude-3-5-sonnet} & Anthropic & 35.46 \\
\texttt{claude-3-5-haiku$\dagger$} & Anthropic & 31.25 \\ 
\texttt{claude-3-sonnet} & Anthropic & 22.63 \\
\texttt{claude-3-haiku} & Anthropic & 18.30 \\ 
\hline
\texttt{llama3-1-405b-instruct} & Meta & 33.19 \\
\texttt{llama3-1-70b-instruct} & Meta & 31.29 \\ 
\texttt{llama3-70b-instruct} & Meta & 27.19 \\ 
\hline 
\texttt{mistral-large-2407$\dagger$} & Mistral & 30.07 \\ 
\hline 
\texttt{command-r-plus$\dagger$} & Cohere & 31.18 \\ 
\bottomrule
\end{tabular} } \caption{Model AST soft scores for the conversational function-calling dataset where chained actions are teacher-forced. $\dagger$ indicates using tool-use formatting.} \label{tab:model_accuracy_scores_by_provider} \end{table}

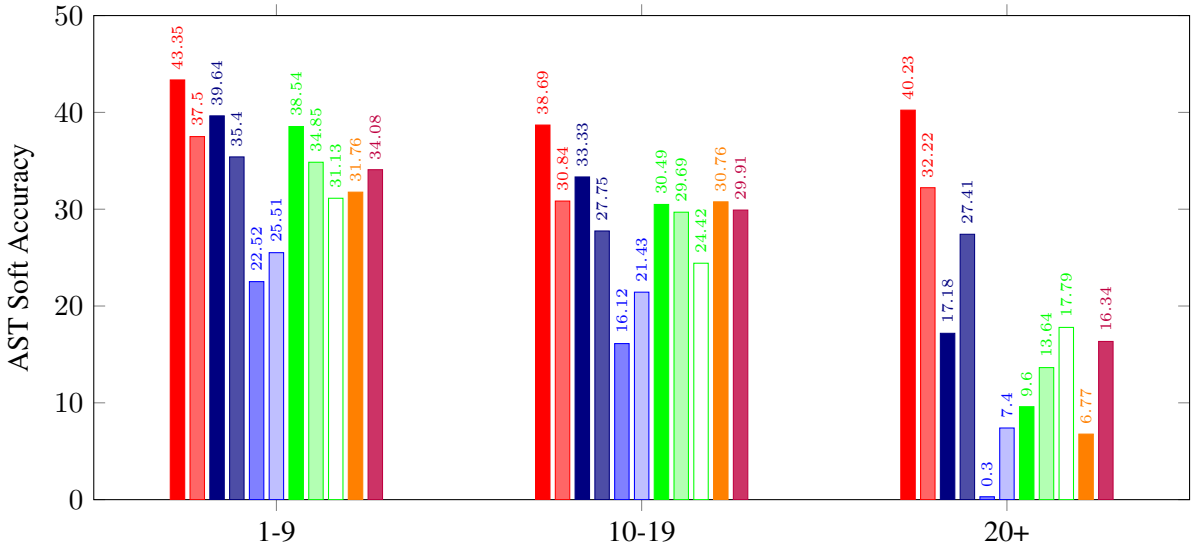
\begin{figure*}[!htb] \makebox[0.9\textwidth][c]{ \begin{tikzpicture} \begin{axis}[ width=\textwidth, height=8cm, ybar, bar width=0.19cm, enlarge x limits=0.25, legend style={at={(0.5,-0.15)}, anchor=north, legend columns=3, /tikz/every even column/.append style={column sep=0.5cm}}, ylabel={AST Soft Accuracy}, symbolic x coords={1-9,10-19,20+}, xtick=data, nodes near coords, nodes near coords align={anchor=west}, every node near coord/.append style={font=\tiny, rotate=90}, ymin=0, ymax=50,
cycle list={ 
{red, fill=red!100}, 
{red, fill=red!60}, 
{darkblue, fill=darkblue!100}, 
{darkblue, fill=darkblue!70}, 
{blue, fill=blue!50}, 
{blue, fill=blue!25}, 
{green, fill=green!100},
{green, fill=green!30}, 
{green, fill=green!0}, 
{orange, fill=orange!100}, 
{purple, fill=purple!80}, 
}, ] 
\addplot coordinates {(1-9,43.35) (10-19,38.69) (20+,40.23)}; 
\addplot coordinates {(1-9,37.50) (10-19,30.84) (20+,32.22)}; 
\addplot coordinates {(1-9, 39.64) (10-19,33.33) (20+,17.18)};   
\addplot coordinates {(1-9,35.40) (10-19,27.75) (20+,27.41)};   
\addplot coordinates {(1-9,22.52) (10-19,16.12) (20+,0.30)};    
\addplot coordinates {(1-9,25.51) (10-19,21.43) (20+,7.40)};    
\addplot coordinates {(1-9,38.54) (10-19,30.49) (20+,9.60)};    
\addplot coordinates {(1-9,34.85) (10-19,29.69) (20+,13.64)};   
\addplot coordinates {(1-9,31.13) (10-19,24.42) (20+,17.79)};   
\addplot coordinates {(1-9,31.76) (10-19,30.76) (20+,6.77)};    
\addplot coordinates {(1-9,34.08) (10-19,29.91) (20+,16.34)};   
\end{axis} 
\end{tikzpicture} } 
\caption{AST soft accuracy for different models across various number of APIs. The dataset has 144 examples with 1-9 APIs, 154 examples of 10-19 APIs and 15 examples of 20+ APIs.} \label{fig:performance_num_apis} \end{figure*}

\begin{figure*}[!htb] \makebox[0.9\textwidth][c]{ \begin{tikzpicture} \begin{axis}[ width=\textwidth, height=8cm, ybar, bar width=0.19cm, enlarge x limits=0.25, legend style={at={(0.5,-0.15)}, anchor=north, legend columns=3, /tikz/every even column/.append style={column sep=0.5cm}}, ylabel={AST Soft Accuracy}, symbolic x coords={1-5,6-10,11+}, xtick=data, nodes near coords, nodes near coords align={anchor=west}, every node near coord/.append style={font=\tiny, rotate=90}, ymin=0, ymax=50, 
cycle list={ 
{red, fill=red!100}, 
{red, fill=red!60}, 
{darkblue, fill=darkblue!100}, 
{darkblue, fill=darkblue!70}, 
{blue, fill=blue!50}, 
{blue, fill=blue!25}, 
{green, fill=green!100},
{green, fill=green!30}, 
{green, fill=green!0}, 
{orange, fill=orange!100}, 
{purple, fill=purple!80}, 
}, ] 
\addplot coordinates {(1-5,44.12) (6-10,37.80) (11+,32.75)}; 
\addplot coordinates {(1-5,35.48) (6-10,31.38) (11+,35.25)}; 
\addplot coordinates {(1-5,33.88) (6-10,41.12) (11+,20.62)};  
\addplot coordinates {(1-5,32.75) (6-10,30.95) (11+,22.22)};  
\addplot coordinates {(1-5,22.72) (6-10,11.92) (11+,16.66)};  
\addplot coordinates {(1-5,26.70) (6-10,16.33) (11+,23.04)};  
\addplot coordinates {(1-5,34.26) (6-10,30.20) (11+,39.42)};  
\addplot coordinates {(1-5,33.12) (6-10,27.68) (11+,35.02)};  
\addplot coordinates {(1-5,31.58) (6-10,24.53) (11+,8.78)};   
\addplot coordinates {(1-5,33.86) (6-10,24.37) (11+,29.64)};  
\addplot coordinates {(1-5,32.48) (6-10,29.35) (11+,30.46)};  
\small\legend{
nova-pro,
nova-micro, 
claude-3-5-sonnet,
claude-3-5-haiku$\dagger$,
claude-3-haiku,
claude-3-sonnet,
llama3-1-405b-instruct,
llama3-1-70b-instruct,
llama3-70b-instruct,
mistral-large-2407-v1:0$\dagger$,
command-r-plus-v1:0$\dagger$
} 
\end{axis} \end{tikzpicture} } \caption{AST soft accuracy for different models across various context lengths. The dataset has 174 examples with 1-5 user turns, 114 examples of 6-10 user turns and 25 examples of 11+ user turns.} \label{fig:performance_context_length} \end{figure*}
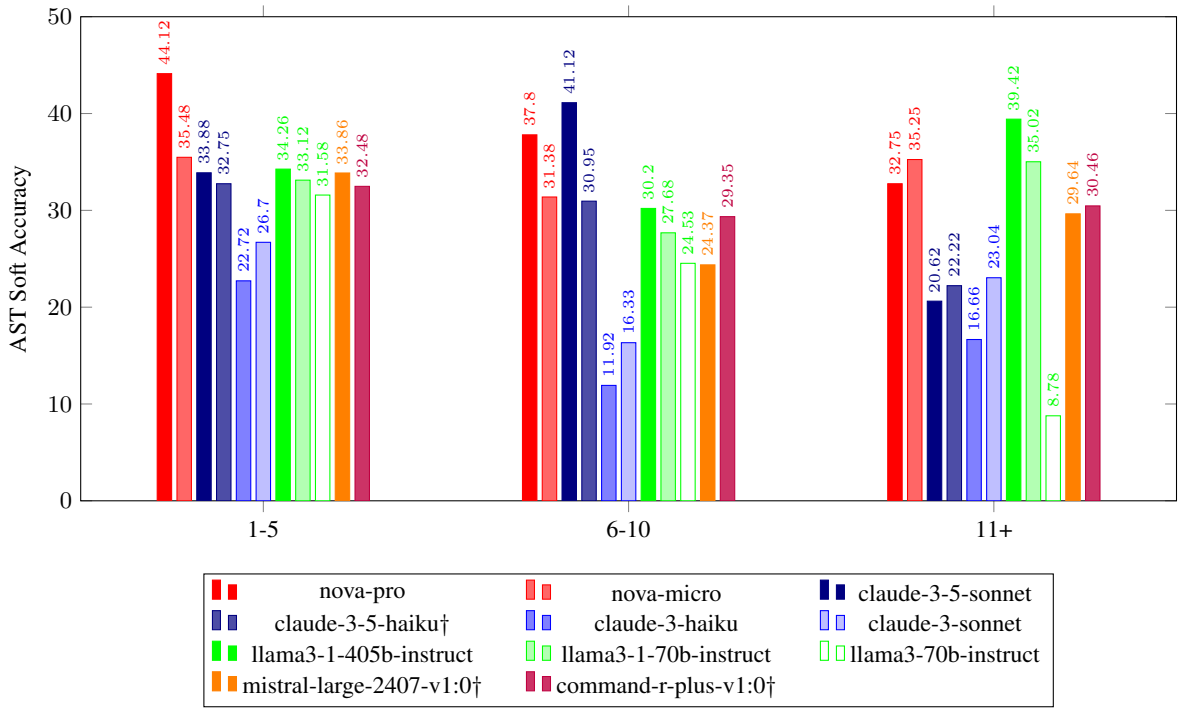    
 
\subsection{Impact of \# of APIs}
We split the data according to the number of APIs included. Figure~\ref{fig:performance_num_apis} shows that while some models such as Nova Pro, Nova Micro, Claude 3.5 Haiku and Mistral large are able to handle 20+ APIs, there is a tendency towards lower performance across the rest of the models when increasing the number of APIs. There are two implications associated with increasing the number of tools: First, since the API schemas go into the prompt, more tools imply longer prompts. Second, more APIs provide higher chances of the model making a mistake since it has more choices to select from. In general, including more tools lead to performance degradation. This applies to Claude, Llama and Command-r-plus models.

\subsection{Context Length Analysis} 
Figure~\ref{fig:performance_context_length} shows results split according to the conversation length, measured by indicating the number of user turns. We observe that while some models such as Nova Pro, Nova Micro, Llama 3.1 405B and 70B  are able to perform  well on long conversations, smaller models such as Haiku v3 struggle to perform well on longer conversations. We also note that Llama 405B appears to perform better on longer conversations. This can be an artifact of off-policy evaluation, where the gold-truth conversation context is leveraged by those models for indirect in-context learning, hence improving their predictions as the context length increases. Claude Sonnet v3.5 and Mistral Large perform well up to 10 user turns, but then tend to degrade for longer conversations. 

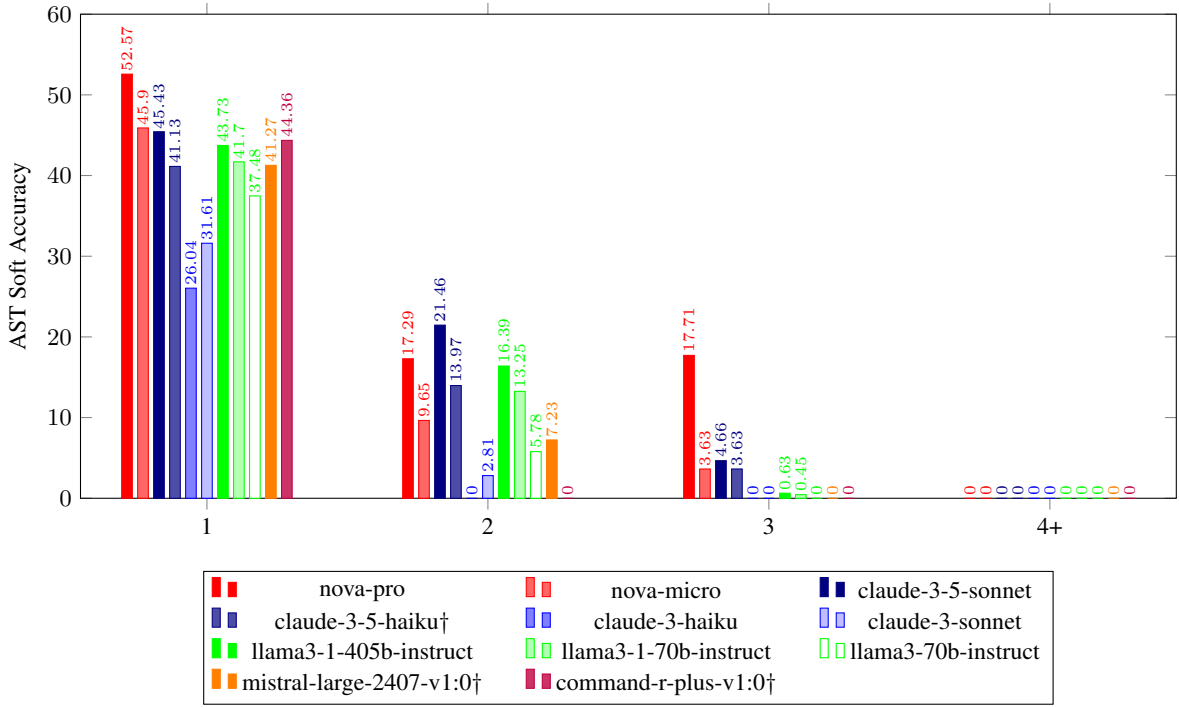
\begin{figure*}[!htb]
\makebox[\textwidth][c]{
\begin{tikzpicture}
\begin{axis}[
    width=\textwidth,
    height=8cm,
    ybar,
    bar width=0.14cm,
    enlarge x limits=0.15,
    legend style={at={(0.5,-0.15)}, anchor=north, legend columns=3, /tikz/every even column/.append style={column sep=0.5cm}},
    ylabel={AST Soft Accuracy},
    symbolic x coords={1,2,3,4+},
    xtick=data,
    nodes near coords,
    nodes near coords align={anchor=west},
    every node near coord/.append style={font=\tiny, rotate=90, inner sep=1pt},
    ymin=0,
    ymax=60,
    x tick label style={font=\small},
    y tick label style={font=\small},
    bar shift auto,
    cycle list={ 
        {red, fill=red!100}, 
        {red, fill=red!60}, 
        {darkblue, fill=darkblue!100}, 
        {darkblue, fill=darkblue!70}, 
        {blue, fill=blue!50}, 
        {blue, fill=blue!25}, 
        {green, fill=green!100},
        {green, fill=green!30}, 
        {green, fill=green!0}, 
        {orange, fill=orange!100}, 
        {purple, fill=purple!80}, 
        }, 
] 
\addplot coordinates {(1,52.57) (2,17.29) (3,17.71) (4+,0)}; 
\addplot coordinates {(1,45.90) (2,9.65) (3,3.63) (4+,0)}; 
\addplot coordinates {(1,45.43) (2,21.46) (3,4.66) (4+,0)}; 
\addplot coordinates {(1,41.13) (2,13.97) (3,3.63) (4+,0.00)}; 
\addplot coordinates {(1,26.04) (2,0) (3,0) (4+,0)}; 
\addplot coordinates {(1,31.61) (2,2.81) (3,0) (4+,0)}; 
\addplot coordinates {(1,43.73) (2,16.39) (3,0.63) (4+,0)}; 
\addplot coordinates {(1,41.70) (2,13.25) (3,0.45) (4+,0)}; 
\addplot coordinates {(1,37.48) (2,5.78) (3,0) (4+,0)}; 
\addplot coordinates {(1,41.27) (2,7.23) (3,0) (4+,0)}; 
\addplot coordinates {(1,44.36) (2,0) (3,0) (4+,0)};    
\small\legend{
nova-pro,
nova-micro, 
claude-3-5-sonnet,
claude-3-5-haiku$\dagger$,
claude-3-haiku,
claude-3-sonnet,
llama3-1-405b-instruct,
llama3-1-70b-instruct,
llama3-70b-instruct,
mistral-large-2407-v1:0$\dagger$,
command-r-plus-v1:0$\dagger$
} 
\end{axis}
\end{tikzpicture}
}
\caption{AST soft accuracy for different models across chain lengths. The dataset has 220 examples with chain length 1, 46 examples with chain length 2,  25 examples with chain length 3, and 22 examples with chain length 4+.}
\label{fig:performance_scores_chain_length}
\end{figure*}

\subsubsection{Chain Length Analysis}

We also investigate the performance on chained action calls, where the model is expected to perform multiple action calls before returning control to the user. For instance, when the model is asked to look up the number of remaining holiday days for an employee, it is expected to retrieve the employee id from the system prior to that call. 
The score of a chain $\tilde{a}_1, \tilde{a}_2, ..., \tilde{a}_J$ of length $J$ actions given a conversation context $c$ is calculated as the product of AST soft accuracies for each of the individual actions: $p(\tilde{a}_1,...,\tilde{a}_J|c)\!=\!\prod_{j=1}^J p(\tilde{a}_j | a_1, ... , a_{j-1}, c)$, 
where $p(\tilde{a}_j | a_1,..., a_{j-1})$ is given by the AST soft accuracy conditioned on gold-truth previous function calls and function call results. Note that we perform teacher-forcing on the action level, where each action is predicted conditioned on the ground-truth previous actions and results.

Figure~\ref{fig:performance_scores_chain_length} shows the results across chains of lengths 1, 2, 3 and 4+. The results indicate that all models have much lower performance than predicting a single action. Claude Sonnet 3.5 demonstrates some ability to predict chains of length 2 with (21.46\%) success, followed by Nova Pro (17.29\%), and Llama 3.1 405B (16.39\%), while the rest of the models perform worse than 15\%. For chains of length 3 and above, almost all models fail in most cases, except for Nova Pro which has 17.71\% AST accuracy.
\begin{figure*}[t]
\small
\begin{subfigure}[b]{0.22\textwidth}
\begin{tikzpicture}[scale=0.55]
  \foreach \y [count=\n] in {
        {0,55,0,0,18}, 
        {77,0,0,0,1}, 
        {5,20,0,0,0},
        {19,4,0,0,7},
        {65,2,0,0,0}, 
    } {
      \ifnum\n<10
            \node[minimum size=6mm, rotate=45, anchor=west] at (\n-0.2, -0.5) {\ifcase\n\or function\or inform\or other\or reject\or seek\_info\fi}; 
      \fi
      \foreach \x [count=\m] in \y {
            \pgfmathsetmacro{\colorval}{min(100,\x/3)} 
        \node[fill=red!\colorval!white, minimum size=6mm, text=black] at (\m,-\n) {\x};
      }
    }
  \node[rotate=90, anchor=south] at (-1.8,-3) {True Dialog Act};
  \foreach \a [count=\i] in {function,inform,other,reject,seek\_info} {
    \node[minimum size=6mm] at (-0.7,-\i) {\a};
  }
\end{tikzpicture}
\caption{mistral-large-2407$\dagger$}
\end{subfigure}
\hfill\hfill\hfill\hfill\hfill\hfill 
\begin{subfigure}[b]{0.22\textwidth}
\begin{tikzpicture}[scale=0.55]
  \foreach \y [count=\n] in {
        {0,26,1,0,8}, 
        {184,0,0,0,0}, 
        {6,18,0,0,1}, 
        {27,1,0,0,0}, 
        {81,0,0,0,0}, 
    } {
      \node[anchor=south] at (3,2) {Predicted Dialog Act};
      \ifnum\n<10
            \node[minimum size=6mm, rotate=45, anchor=west] at (\n-0.2, -0.5) {\ifcase\n\or function\or inform\or other\or reject\or seek\_info\fi}; 
      \fi
      \foreach \x [count=\m] in \y {
        \pgfmathsetmacro{\colorval}{min(100,\x/3)} 
        \node[fill=red!\colorval!white, minimum size=6mm, text=black   ] at (\m,-\n) {\x};
      }
    }
\end{tikzpicture}
\caption{Llama 3.1 70B}
\end{subfigure}
\begin{subfigure}[b]{0.22\textwidth}
\begin{tikzpicture}[scale=0.55]
  \foreach \y [count=\n] in {
        {0,29,0,1,29}, 
        {51,0,0,0,0}, 
        {5,12,0,1,7}, 
        {14,4,0,0,3}, 
        {29,1,0,0,0}, 
    } {
      \ifnum\n<10
            \node[minimum size=6mm, rotate=45, anchor=west] at (\n-0.2, -0.5) {\ifcase\n\or function\or inform\or other\or reject\or seek\_info\fi}; 
      \fi
      \foreach \x [count=\m] in \y {
      \pgfmathsetmacro{\colorval}{min(100,\x/3)} 
        \node[fill=red!\colorval!white, minimum size=6mm, text=black] at (\m,-\n) {\x};
      }
    }
    
\end{tikzpicture}
\caption{claude-sonnet-3-5$\dagger$}
\end{subfigure}
\begin{subfigure}[b]{0.22\textwidth}
\begin{tikzpicture}[scale=0.55]
  \foreach \y [count=\n] in {
        {0,68,0,1,24}, 
        {43,0,0,0,1}, 
        {1,23,0,0,1}, 
        {12,6,0,0,5}, 
        {50,5,0,0,0}, 
    } {
      \ifnum\n<10
            \node[minimum size=6mm, rotate=45, anchor=west] at (\n-0.2, -0.5) {\ifcase\n\or function\or inform\or other\or reject\or seek\_info\fi}; 
      \fi
      \foreach \x [count=\m] in \y {
      \pgfmathsetmacro{\colorval}{min(100,\x/3)} 
        \node[fill=red!\colorval!white, minimum size=6mm, text=black] at (\m,-\n) {\x};
      }
    }
\end{tikzpicture}
\caption{command-r-plus$\dagger$}
\end{subfigure}
\caption{Confusion matrix for dialog acts for (a) mistral-large-2407  (b) llama3-1-70b-instruct, (c) claude-sonnet-3-5, and (d) command-r-plus. The rows correspond to true labels and the columns to the predicted labels. We are not showing the diagonal values which are correct matches. $\dagger$ indicates using tool-use formatting.}
\label{fig:confusion_matrix_heatmap_sonnet35}
\end{figure*}
\subsubsection{Response Quality via Dialog Acts}\label{sec:da_results}
To help understand where and how models fail, we compare model responses using coarse-grained dialog acts: inform user, seek information, reject request, function call, and other. We use an LLM classifier based on Claude Haiku v3 to classify the dialog acts of the model responses into one of these five categories. We report the classification performance on the ground-truth dataset in Table~\ref{tab:da_classifier_metrics}. The classifier predicts one or more dialog act labels for the model response, where the ground-truth annotations may include one or more dialog act labels. We note that the classifier has a high recall but low precision. In our reporting, we declare a match if at least one of the predicted dialog acts matches any of the ground-truth labels. The classification prompt is given in Table~\ref{tab:da_prompt} in Appendix~\ref{sec:app_hallucination}.

\begin{table}[h]
\small
\centering 
\begin{tabular}{lc}
\toprule
\textsc{Metric} & \textsc{Score (\%)} \\ 
\hline
\textsc{Multi-label Precision} & 67.2 \\
\textsc{Multi-label Recall} & 90.3 \\ 
\textsc{Multi-label F1 Score} & 77.0 \\ 
\hline
\textsc{Single-label Accuracy}  & 92.8\\
\bottomrule
\end{tabular} \caption{Dialog act classifier performance on gold-truth conversations. The classifier detects multiple labels which are compared to the multi-label ground-truth labels. For Single-label accuracy we calculate how often at least one predicted label matches one of the ground-truth labels.} \label{tab:da_classifier_metrics} \end{table}

In terms of dialog act accuracy, Claude Sonnet v3.5 (FC) is the best performing with 73.13\% accuracy, followed closely by Nova Pro (69.68\%), Haiku 3.5 (67.12\%), Nova Micro (66.21\%) and command-r-plus (64.86\%). Llama models perform worse than the rest. We further analyze the confusion matrix for the dialog acts in Figure~\ref{fig:confusion_matrix_heatmap_sonnet35}. We omit the correct matches on the diagonals and focus on the mismatches. In general we see a tendency to confuse inform and seek information acts with function calls, indicating that the models are falsely triggering function calls, although to different extents depending on the model. This is showing to a large extent in the Llama 3.1 70B model. We also observe under-triggering of function calls where the models tend to inform the user or seek information rather than calling a function. 

\begin{table}[t] \centering 
\resizebox{\columnwidth}{!}{
\begin{tabular}{lc} 
\toprule
\textsc{Model} & \textsc{DA Accuracy (\%) $\uparrow$} \\ 
\hline
\texttt{nova-pro$\dagger$} &  69.68  \\
\texttt{nova-micro$\dagger$} & 66.21  \\
 \hline 
\texttt{claude-3-5-sonnet$\dagger$} & 73.15 \\
\texttt{claude-3-5-haiku$\dagger$} & 67.12 \\ 
\texttt{claude-3-sonnet} &  61.54 \\
\texttt{claude-3-haiku} & 63.80 \\ 
\hline
\texttt{llama3-1-70b-instruct} & 50.98 \\
\texttt{llama3-70b-instruct} & 56.26 \\
\hline 
\texttt{mistral-large-2407$\dagger$} & 61.09 \\
\hline 
\texttt{command-r-plus$\dagger$} & 64.86 \\ 
\bottomrule
\end{tabular} 
}
\caption{Dialog acts accuracy for different models. $\dagger$~indicates using tool-use formatting.} \label{tab:dialog_acts_accuracy} \end{table}\vspace{-1mm}
\subsection{Parameter Hallucination Analysis}\label{sec:param_hallucinations}
 Table~\ref{tab:hallucination_scores} shows hallucination results for all models. We report  parameter validity rate for the ground truth as a reference. We observe that Claude v3.5 Sonnet has the highest validity rate (72.4\%), followed by Nova Pro (68.8\%), Haiku 3.5 (66.1\%), Command-R Plus (63.7\%) and Llama 3.1 405B (62.7\%). 
 Claude Sonnet v3 and Haiku v3 have much lower scores.   

\begin{table}[htb!] 
\centering 
\resizebox{\columnwidth}{!}{
\begin{tabular}{lc}
\toprule
& \textsc{Valid}\\
\textsc{Model}  & \textsc{Parameters ($\uparrow$)}\\\hline
\texttt{Ground Truth}&$83.1\%$\\\hline
\texttt{Nova Pro} &  $68.8\%$  \\
 \texttt{Nova Micro} &  $57.9\%$ \\
 \hline 
\texttt{claude-3-5-sonnet}&$72.4\%$\\
\texttt{claude-3-5-haiku$\dagger$}&$66.1\%$\\
\texttt{claude-3-sonnet}&$40.7\%$\\
\texttt{claude-3-haiku}&$36.3\%$\\
\hline
\texttt{llama3-1-405b-instruct}&$62.7\%$\\
\texttt{llama3-1-70b-instruct}&$59.2\%$\\
\texttt{llama3-70b-instruct}&$51.4\%$\\\hline
\texttt{mistral-large-2407$\dagger$}&$51.1\%$\\\hline
\texttt{command-r-plus$\dagger$}&$63.7\%$\\
\bottomrule
\end{tabular} 
}
\caption{Parameter validity rate for different models. \textsc{Valid Parameters} is the percentage of valid predicted parameters for valid action calls as calculated in Section~\ref{sec:param_valid}. $\dagger$ indicates using tool-use formatting.}
\label{tab:hallucination_scores} 
\end{table}
\vspace{-1mm}
\section{Conclusion}
We introduced a conversational benchmark for function calling spanning multiple complexities natural to dialog. The dataset targets the evaluation of function calling and response quality. The benchmark includes over 80 API tools and tests function-calling up to 20+ APIs under different conversation lengths. We evaluated top LLMs on our proposed benchmark in an off-policy turn-level fashion, where we set the context to the ground-truth conversation up to the current turn. The context is composed of user and agent turns as well as function calls and results. Overall, the models performance on this benchmark is relatively low, suggesting that LLMs need further work to improve conversational function-calling capabilities. The benchmark can be used to assess model performance across varying conversations length, number of tools and action chain lengths. 
\section{Limitations \& Risks}
Turn-level evaluation is a fine-grained evaluation  approach that allows detailed analysis of how and when predictions deviate from the ground truth. However, performing turn-level evaluation is done in an off-policy fashion, where the gold truth is fixed as context. This does not correspond directly to inference behavior, where the model uses on-policy predictions for agent responses as context. For instance, using off-policy trajectories can induce a form of in-context learning that help later predictions conform to the generation style of the gold truth, or even learn action-calling dependencies from ground-truth context, causing an artificial inflation in model performance for later turns.
Another deviation from inference-time behavior is regarding function call execution. In this work, we do not execute function calls. All function call results are looked up from gold truth turns.

The dialog act accuracy and parameter validity rate are both calculated leveraging LLMs to classify or judge the outcome. These LLMs demonstrate high recall for dialog act classification and relatively high parameter validity rate for the ground-truth annotations. However, these models can still make mistakes which can obscure the true model performance. 

\paragraph{Data Statement} We provide a data statement in adherence with the ACL code of conduct and recommendations laid out in \cite{bender-friedman-2018-data}. The dataset was collected by linguists who are professional English speakers hired through a vendor and they were remunerated above industry standard rates. The conversations were simulated by human annotators, where the same annotator played both the user role and the agent roles, and created the function calls.  All conversation parts were collected from scratch except for function call results, which were simulated using an internally trained model. The dataset went through quality assurance process done internally, where conversations were edited as needed to fix any observed errors.

\section{Acknowledgments}
We would like to thank Daniele Bonadiman and Sailik Sengupta for their contributions in creating the APIs, I-Hsuan Chen for support with the  dialog act annotation collection, and Emily Moeng for managing data collection through vendors.
\bibliography{custom}

\appendix

\begin{table*}[t]
\centering
\begin{tcolorbox}[
    enhanced,
    width=\textwidth,
    colback=gray!10,
    colframe=gray!50,
    arc=0mm,
    boxrule=0.5pt,
    top=3mm,
    bottom=3mm,
    left=3mm,
    right=3mm
]
\begin{tabular}{p{0.95\textwidth}}
\toprule
\textbf{System Prompt and Instructions for Dialog Act Classification} \\
\midrule
\begin{minipage}[t]{\linewidth}
\footnotesize
\begin{verbatim}

SYSTEM PROMPT:

"""
output the labels only. Do not output anything else. 
If more than one label is predicted, separate them using comma (,).
"""

USER PROMPT:

"""
Select one or more of the following labels as the dialog act for the following text.
If the turns include only function or API calls, use the agent_function_calling label only.
Only use the other labels if the turns include text that is not function-calling. function-calls
may co-occur with addition text informing user or seeking information from the user or rejecting 
the user request.
labels are: agent_seek_information, agent_other, agent_reject_request, agent_inform_user,
agent_function_calling
text: {prediction}
conversation history: {context}"
"""
\end{verbatim}
\end{minipage} \\
\bottomrule
\end{tabular}
\end{tcolorbox}
\caption{The system prompt and the instructions used for dialog act classification.}
\label{tab:da_prompt}
\vspace{1ex}
\captionsetup{justification=raggedright,singlelinecheck=false}
\end{table*}

\begin{table*}[t]
\centering
\begin{tcolorbox}[
    enhanced,
    width=\textwidth,
    colback=gray!10,
    colframe=gray!50,
    arc=0mm,
    boxrule=0.5pt,
    top=3mm,
    bottom=3mm,
    left=3mm,
    right=3mm
]
\begin{tabular}{p{0.95\textwidth}}
\toprule
\textbf{Function-calling Evaluation Prompt} \\
\midrule
\begin{minipage}[t]{\linewidth}
\footnotesize
\begin{verbatim}

SYSTEM PROMPT:

"""
You are an expert in composing functions. You are given a conversation and a set of possible functions. 
Based on the conversation, you will need to make one function/tool call to achieve the purpose. 
If you need to call multiple function calls to achieve the purpose, output the first function
call in the chain only.
If none of the functions can be used, respond with an appropriate message to the user, 
such as clarification, confirmation, eliciting information, etc. If the given question lacks 
the parameters required by the function, you can ask the user for it. You should only return 
the function call in tools call sections.

USER PROMPT (only used when prompting models directly):

"""
Here is a list of functions in JSON format that you can invoke. 
{language_specific_hint}\n\n{functions}\n\n
If you decide to invoke any of the function(s), put it in the format of
[func_name1(params_name1=params_value1, params_name2=params_value2...), func_name2(params)]\n
You SHOULD NOT include any other information in the response.
{conversation}
"""
\end{verbatim}
\end{minipage} \\
\bottomrule
\end{tabular}
\end{tcolorbox}
\caption{The system and user prompts used to generate predictions for the function-calling benchmark. The user prompt is only included for the case of direct prompting (i.e. when the tool use API is not used). These runs are indicated by dropping the "FC" suffix in the results tables.}
\label{tab:func_prompt}
\vspace{1ex}
\captionsetup{justification=raggedright,singlelinecheck=false}
\end{table*}

\begin{table*}[t]
\centering
\begin{tcolorbox}[
    enhanced,
    width=\textwidth,
    colback=gray!10,
    colframe=gray!50,
    arc=0mm,
    boxrule=0.5pt,
    top=3mm,
    bottom=3mm,
    left=3mm,
    right=3mm
]
\begin{tabular}{p{0.95\textwidth}}
\toprule
\textbf{Response Quality Evaluation Prompt} \\
\midrule
\begin{minipage}[t]{\linewidth}
\footnotesize
\begin{verbatim}

SYSTEM PROMPT:

"""
You are an agent tasked with assisting a user to solve a problem. You have access to a set of 
functions that you can use. 
You are given a conversation and a set of possible functions. Based on the conversation, 
you will need to either make one function/tool call or to respond to the user to achieve the purpose. 
If you need to make multiple function calls to achieve the purpose, output the first function
call in the chain only. 
If none of the functions can be used, respond with an appropriate message to the user, such as
clarification, confirmation, 
eliciting information, etc. If the given question lacks the parameters required by the function, 
you can ask the user for it.

USER PROMPT (only used when prompting models directly):

"""
Here is a list of functions in JSON format that you can invoke. 
{language_specific_hint}\n\n{functions}\n\n
If you decide to invoke any of the function(s), put it in the format of 
[func_name1(params_name1=params_value1, params_name2=params_value2...), func_name2(params)]\n
{conversation}
"""
\end{verbatim}
\end{minipage} \\
\bottomrule
\end{tabular}
\end{tcolorbox}
\caption{The system and user prompts used to generate predictions for the response quality benchmark. The user prompt is only included for the case of direct prompting (i.e. when the tool use API is not used). These runs are indicated by dropping the "FC" suffix in the results tables.}
\label{tab:response_prompt}
\vspace{1ex}
\captionsetup{justification=raggedright,singlelinecheck=false}
\end{table*}

\section{Prompts}\label{sec:app_hallucination}
Table~\ref{tab:da_prompt} provides the prompts used for dialog act classification using Claude Haiku v3. We also include the system and user prompts used to generate predictions for evaluating function calling (Section~\ref{sec:func_call_eval_results}) and response quality (Section~\ref{sec:da_results}) in Table~\ref{tab:func_prompt} and Table~\ref{tab:response_prompt}, respectively. Table~\ref{tab:hallucination_prompt} shows the prompt we used to evaluate parameter hallucinations using \texttt{gpt-4o-mini} (with temperature $0$) in Section~\ref{sec:param_hallucinations}.
\begin{table*}[t]
\centering
\begin{tcolorbox}[
    enhanced,
    width=\textwidth,
    colback=gray!10,
    colframe=gray!50,
    arc=0mm,
    boxrule=0.5pt,
    top=3mm,
    bottom=3mm,
    left=3mm,
    right=3mm
]
\begin{tabular}{p{0.95\textwidth}}
\toprule
\textbf{Prompt for Parameter Hallucination Detection} \\
\midrule
\begin{minipage}[t]{\linewidth}
\footnotesize
\begin{verbatim}
"""
<guidelines>
- The system is about to execute an action or a series of actions shown 
  below inside <actions></actions> tags. Each predicted action is called 
  with the action name, parameter name and parameter values, as shown in 
  the action schemas inside <schemas></schemas> tags. 
- You have to evaluate each parameter value in the predicted action calls 
  and determine whether it was hallucinated.
- Each parameter value must be EXPLICITLY mentioned in the conversation 
  history. If not, the value is hallucinated. For example if an airport 
  code is not explicitly mentioned in the context and the system uses it, 
  it is considered hallucination. Note that certain actions might require 
  the system to generate text (such as an email or an article). In these 
  cases, evaluate the content of the generated text for hallucination.
- First think before detecting hallucinated parameters inside 
  <thinking></thinking> tags and add the rationale in the "rationale" field.
- The output should have the following JSON format:: 
  {"rationale": "", "hallucinated_parameters": [[], [], ...]}. The 
  "hallucinated_parameters" field should contain a list that has one list 
  of hallucinated parameter names for each predicted action in the order 
  as they appear.
</guidelines>

This is the action call you have to evaluate:

Conversation History:
<conversation_history>
{conversation_history_block}
</conversation_history>

Predicted Actions:
<actions>
{predicted_actions}
</actions>

Action Schemas:
<schemas>
{action_schemas}
</schemas>

Evaluate the parameters of the predicted actions in <actions></actions> 
tags. Return a JSON as described in the <guidelines>.
"""
\end{verbatim}
\end{minipage} \\
\bottomrule
\end{tabular}
\end{tcolorbox}
\caption{Prompt for detecting parameter hallucination in predicted actions}
\label{tab:hallucination_prompt}
\vspace{1ex}
\captionsetup{justification=raggedright,singlelinecheck=false}
\end{table*}
\end{document}